\documentclass[letterpaper]{article} 
\usepackage{aaai25}  
\usepackage{times}  
\usepackage{helvet}  
\usepackage{courier}  
\usepackage[hyphens]{url}  
\usepackage{graphicx} 
\usepackage{natbib}  
\usepackage{caption} 
\usepackage{algorithm}
\usepackage{algpseudocode}
\usepackage{multirow}
\usepackage{pifont}
\usepackage{multicol}
\usepackage{enumitem} 
\usepackage{booktabs}       

\usepackage{newfloat}
\usepackage{listings}
\DeclareCaptionStyle{ruled}{labelfont=normalfont,labelsep=colon,strut=off} 
\floatstyle{ruled}
\newfloat{listing}{tb}{lst}{}
\floatname{listing}{Listing}
\usepackage{amsfonts,amsmath,amsthm,amssymb}

\title{EventAug: Multifaceted Spatio-Temporal Data Augmentation\\Methods for Event-based Learning}
\author {
   Yukun Tian\textsuperscript{\rm 1},
    Hao Chen\textsuperscript{\rm 1}
\thanks{Correspongding Author},
    Yongjian Deng\textsuperscript{\rm 2},
    Feihong Shen\textsuperscript{\rm 1},
    Kepan Liu\textsuperscript{\rm 3},
    Wei You\textsuperscript{\rm 4},
    Ziyang Zhang\textsuperscript{\rm 4}
}
\affiliations{
    \textsuperscript{\rm 1}School of Computer Science and Engineering, Southeast University,\\
    \textsuperscript{\rm 2}College of Computer Science, Beijing University of Technology,\\
      \textsuperscript{\rm 3}Beijing Institute of Technology,\\
    \textsuperscript{\rm 4}Huawei Technologies Co., Ltd,\\

   haochen303@seu.edu.cn
}

\usepackage{bibentry}

\begin{document}

\maketitle

\begin{abstract}
The event camera has demonstrated significant success across a wide range of areas due to its low time latency and high dynamic range. However, the community faces challenges such as data deficiency and limited diversity, often resulting in over-fitting and inadequate feature learning. Notably, the exploration of data augmentation techniques in the event community remains scarce.
This work aims to address this gap by introducing a systematic augmentation scheme named EventAug to enrich spatial-temporal diversity. In particular, we first propose Multi-scale Temporal Integration (MSTI) to diversify the motion speed of objects, then introduce Spatial-salient Event Mask (SSEM) and Temporal-salient Event Mask (TSEM) to enrich object variants. 
Our EventAug can facilitate models learning with richer motion patterns, object variants and local spatio-temporal relations, thus improving model robustness to varied moving speeds, occlusions, and action disruptions. 
Experiment results show that our augmentation method consistently yields significant improvements across different tasks and backbones (\textit{e.g.,} a 4.87\% accuracy gain on DVS128 Gesture). Our code will be publicly available for this community.
\end{abstract}

%
\section{Introduction}\label{introduction}
\begin{figure}[h]
	\centering
	\begin{minipage}[b]{0.45\textwidth}
		\centering
		\includegraphics[width=\textwidth]{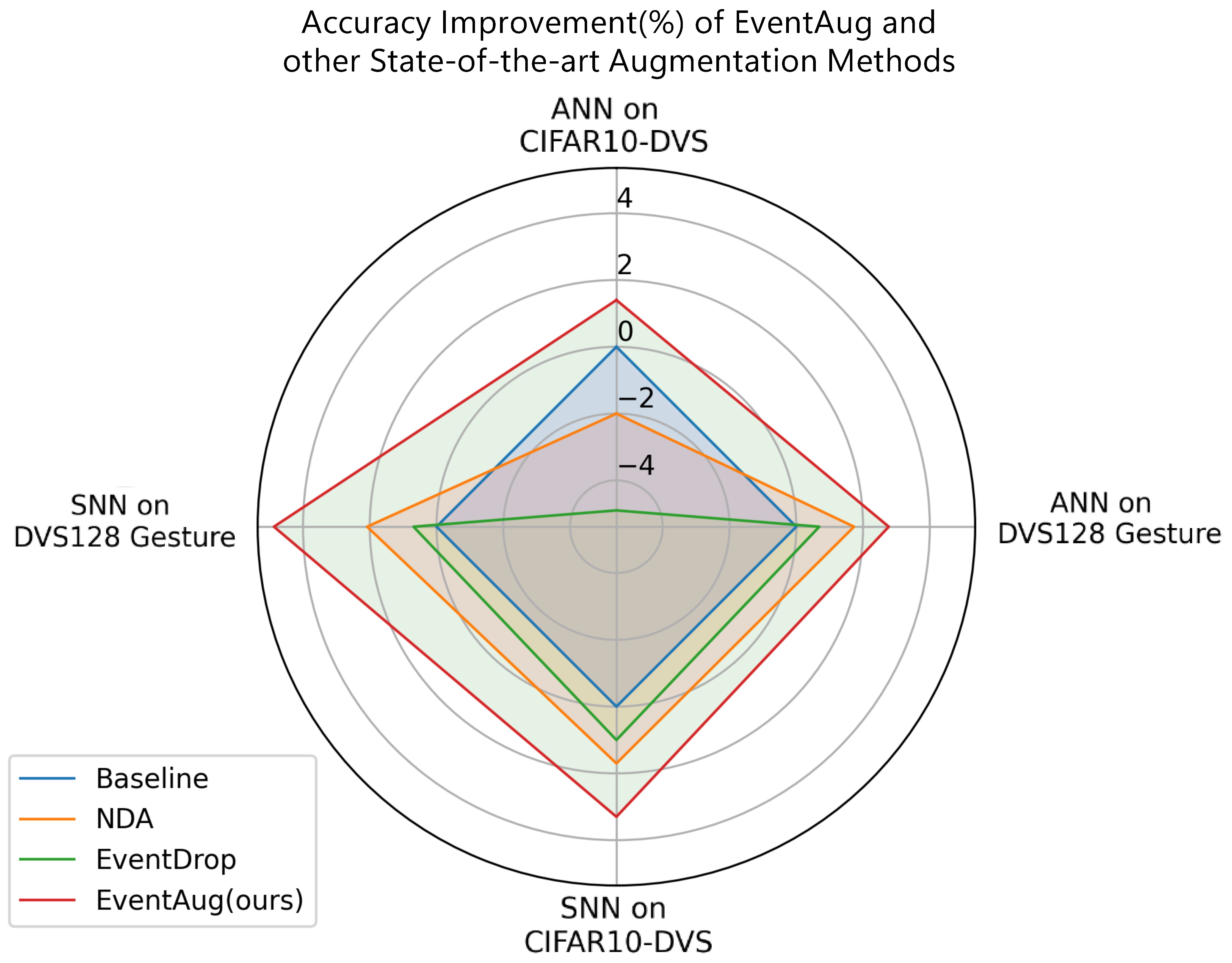} 
		\caption{Comparison of our EventAug and other state-of-the-art augmentation methods on different tasks and kinds of backbones.}
		\label{lidar}
	\end{minipage}
	\hfill

\end{figure}
\begin{figure}
    \centering
    \includegraphics[width=1.00\linewidth]{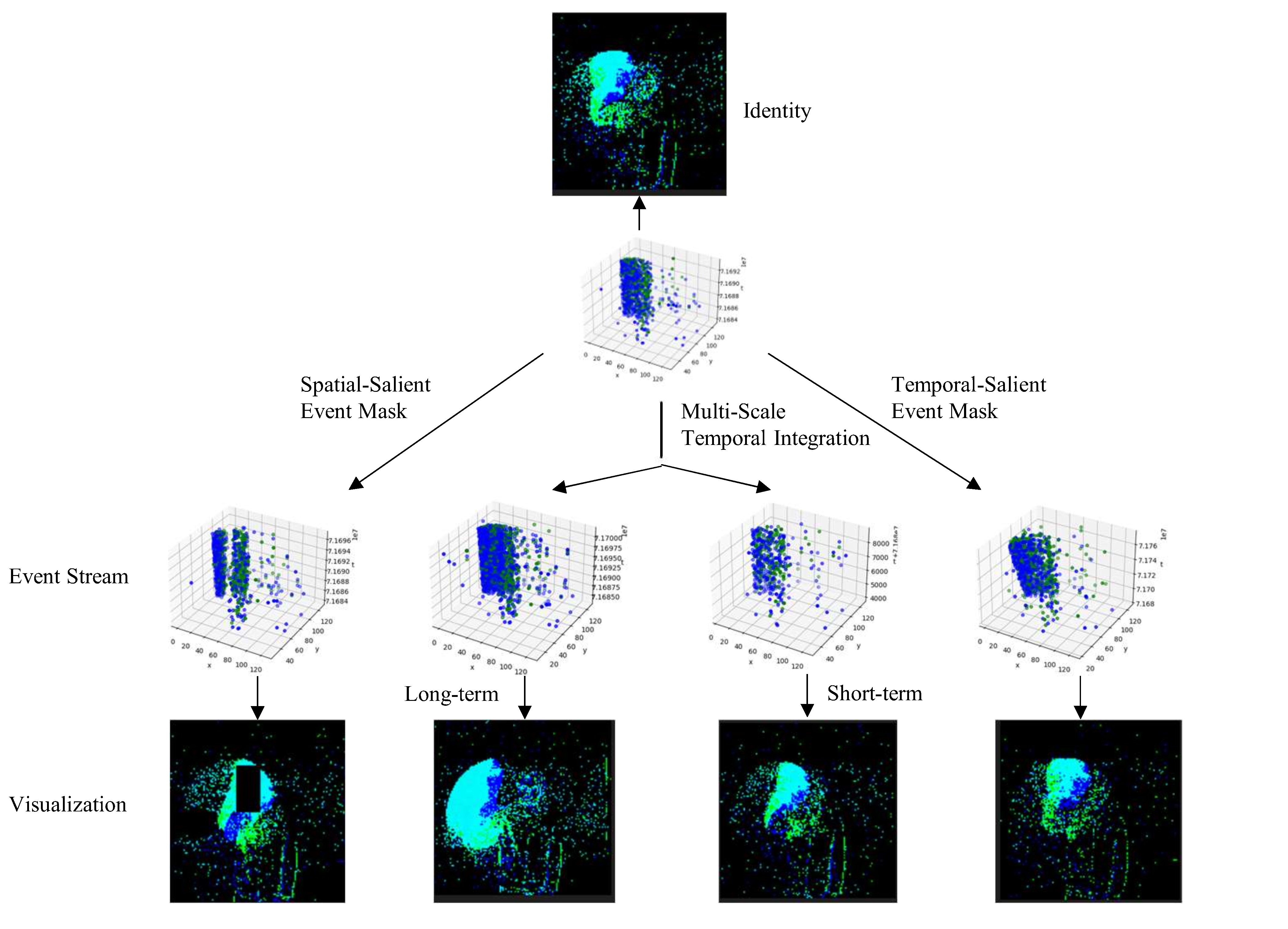}
    \caption{An example of augmented events with our EventAug, including the visualization of the original and augmented event stream and event frame. Our methods greatly enhance the diversity of the original dataset, which improves the model’s generalization abilities.}
    \label{vis}
\end{figure}
\begin{figure*}[h]
	\centering
	\includegraphics[width=0.9\linewidth]{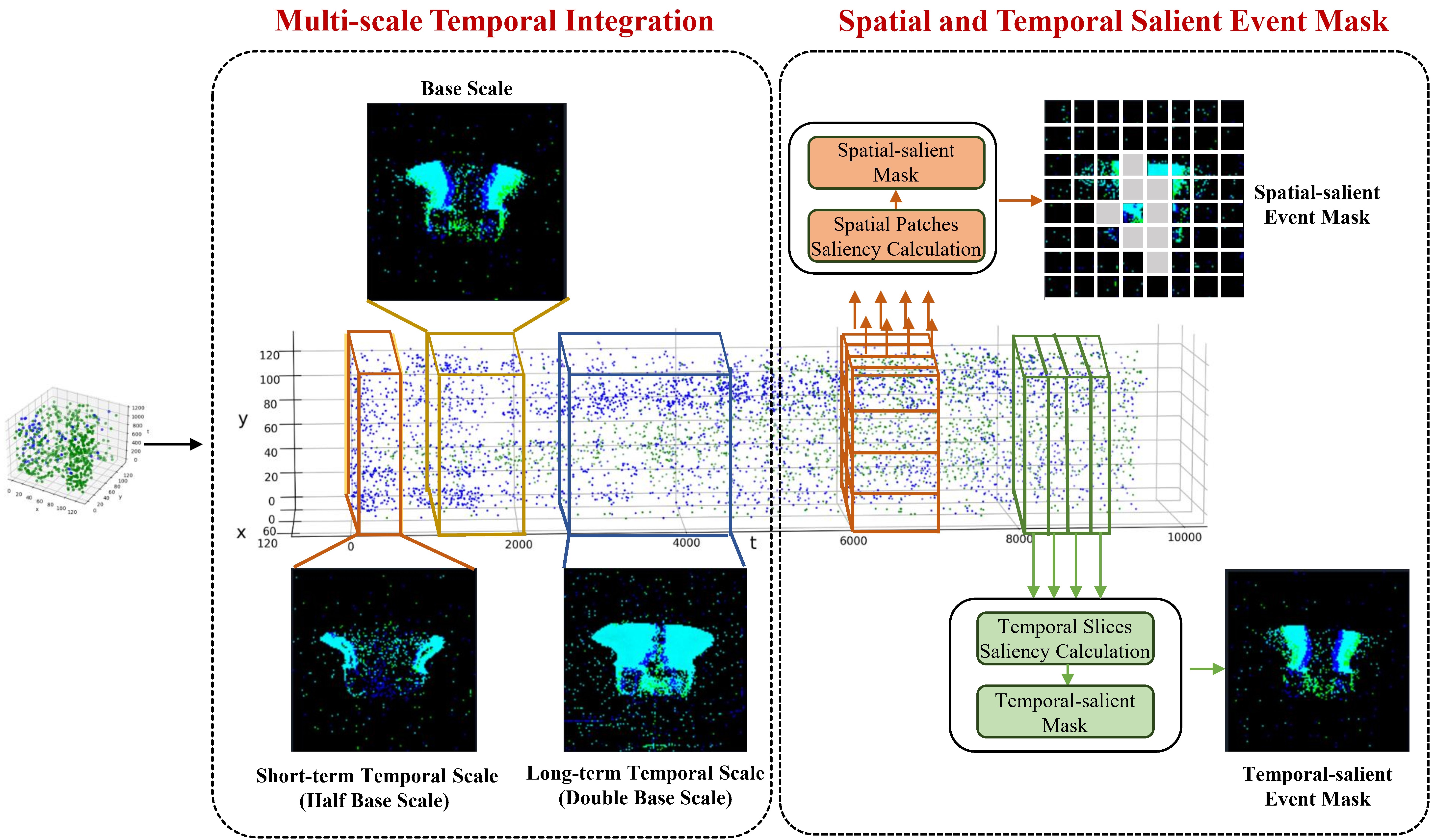}
	\caption{Illustration of our EventAug methods. The left is Multi-scale Temporal Integration. Since frames generated by short-term and long-term temporal scale reveal different motion patterns. Therefore, by applying a multi-scale integration strategy, we enable the model to better learn motion information together with edge feature. The right is Spatial and Temporal Salient Event Mask. Guided by the saliency information, we selectively mask event in salient spatial patches and salient temporal slices.}
	\label{overview}
\end{figure*}
Event camera, which is also known as Dynamic Vision Sensors(DVS)\cite{eventcamera}, is a new kind of bio-inspired device. Unlike conventional RGB frame cameras, event camera only focuses on the changes but not the absolute value of brightness, thus it has several unique features, including low-latency, low energy consumption, and extremely high dynamic range. These advantages make event camera a powerful tool in research areas like classification\cite{classification}, depth estimation\cite{depthestimation}, flow estimation\cite{flowestimation} and motion segmentation\cite{motionsegmentation}. These advantages greatly stimulate the research in event-based learning area. \\
\indent Existing works in event-based learning community mainly focus on backbone design and task-specific network building. Representative works include the Group Event Transformer \cite{GET}, Spikepoint \cite{spikepoint}, video deraining \cite{derain}, and motion deblurring \cite{motiondeblur}. However, event data is sparse and limited in quantity, leading to annotation difficulties and a scarcity of high-quality labeled data. This results in over-fitting and insufficient feature extraction, constraining the performance and application of event data. Spike Neural Networks (SNNs), considered more suitable for handling sparse event data, require greater data diversity due to challenges in optimization and training. Hence, developing methods to reduce over-fitting and enhance model performance across various tasks is a critical research priority.


Data augmentation, which has been proved to be an effective way to improve the generalization ability of models for RGB images, is a practical method to solve the problems above. Yet, rare research focuses on data augmentation in event community. Currently, there are two main kinds of event augmentation strategies. \textit{(i)} Directly transferring the conventional data augmentation methods for images to the event frames, for example, applying the geometry transformation for RGB images on event data, like flipping, rolling and rotation \cite{NDA}. These augmentation strategies apply the paradigms designed for RGB modality and ignore the sparsity and temporal characteristics of event data, resulting in limited augmentation effects and a failure to generalize across complex real-world situations like varied moving speed, occlusion, and action disruption. 
\textit{(ii)}Utilizing the temporal information of the event data in a coarse manner. Representative work includes randomly dropping event data \cite{EventDrop} and mixing up two event streams with a three-dimensional mixing strategies generated by a random Gaussian Mixture Model \cite{EventMix}. These methods rely heavily on randomness and prior assumptions(\textit{e.g.,}Gaussian distribution, uniform distribution) which ignores the uneven spatial and temporal distribution of event data, for example, when augmentation is applied to irrelevant spatial or temporal area, the result will be considerably poor and inefficient. Thus the augmentation performance is very limited and lacks generality. Moreover, since they do not reveal the rich motion and object information inside event data, they only make elementary use of spatio-temporal information in event data, making it incapable of effectively enhancing the spatio-temporal diversity of the dataset. 



Given the problems of the existing methods described above, we believe that a customized event data augmentation approach should be designed to enhance training data diversity efficiently by considering both the sparsity characteristics of event data and their uneven spatio-temporal distribution. By utilizing such augmentation techniques, we can enhance the diversity of the datasets and improve the model’s generalization capabilities (Figure \ref{lidar},\ref{vis}).

To this end, we propose the EventAug (Figure \ref{overview}), which contains three novel spatio-temporal augmentation methods for event data such as \textit{Multi-scale Temporal Integration(MSTI)}, \textit{Spatial-salient Event Mask(SSEM)} and \textit{Temporal-salient Event Mask(TSEM)}. \textbf{They are proposed to fully utilize the rich spatio-temporal information inside event data and enhance the diversity of training samples considering comprehensively the sparse and uneven spatio-temporal distribution properties of event data.} 

For MSTI, since motion speed determines the completeness of motion cues and the clarity of object boundaries within an event frame, varied moving speeds can capture different temporal and spatial patterns. Therefore, we enrich the diversity of motion speeds by adjusting integration scale to form a multi-scale temporal integration. \textbf{Precisely, by applying a multi-scale augmentation strategy, we actually simulated different motion patterns, which enable us to generate samples under diverse motion scenarios utilizing a single event-based sample. Features such as edges, textures, and motion also change with the integration scale, thereby enhancing diversity and making the network more robust to diverse motion patterns.} Also, MSTI can boost the generalization ability across sensor noise since frames generated from different integration scales possess diverse noise levels.

\textbf{For MSTI and TSEM, we employ spatial and temporal mask to diversify local spatial and temporal correlations, thus enriching the high-level semantics of event data and inducing the models to learn more spatio-temporal relations.} Specifically, to address the uneven spatial-temporal distribution of event data, we propose a fast, training-free spatial saliency and temporal saliency calculation method to obtain saliency with low computational cost. With the guidance of saliency information, our augmentation is highly effective, stable, and adaptive to different event datasets. Comparing to former methods with strong prior assumptions \cite{EventDrop,EventMix}, we only apply augmentation in salient spatial and temporal regions and the amplitude of augmentation is defined adaptively $w.r.t$ the strength of saliency. Moreover, the design philosophy of MSTI and TSEM make them able to greatly improve the robustness towards occlusion and motion disruption, thus significantly improving generalization ability of models on downstream tasks on complex real-world scenarios. 

In summary, our main contributions are as follows:

	(1) We propose the Multi-scale Temporal Integration (MSTI) technique to enhance the diversity of motion speeds. MSTI allows an event model to learn additional motion cues and spatial features. This provides the model with enhanced generalization capabilities across different scenarios involving objects moving at different speeds.
	
	(2) We introduce two methods, namely Spatial-salient Event Mask (SSEM) and Temporal-salient Event Mask (TSEM), to diversify local correlations using fast spatial and temporal saliency guidance. These methods address the uneven spatial and temporal distribution of event data, significantly improve the diversity of local spatio-temporal correlations, and enhance robustness to occlusion and motion disruption in complex scenarios.
	
	(3) Experimental results on both ANN and SNN networks demonstrate that our proposed methods comprehensively enhance spatio-temporal diversity with high efficiency. These improvements lead to significant enhancements in accuracy and generalization ability.
\section{Related work}\label{related work}
\subsection{Event-based Learning}
Recently, event-based learning has become a popular research area due to the development of Dynamic Vision Sensor(DVS) and neuromorphic computing \cite{nc1,nc2,nc3}. Existing work in event-based learning community mainly focuses on backbone design and task-specific network building. Some representative works include Group Event Transformer \cite{GET}, Spikepoint \cite{spikepoint}, video deraining \cite{derain} and motion deblurring \cite{motiondeblur}. However, the limited amount of event data and the large variation between event datasets restrict the performance of the network and finally lead to poor model generalization. In order to overcome these difficulties and further utilize the power of event data, many learning strategies have been proposed, such as unsupervised learning \cite{derain}, self-supervised learning \cite{tlearn2}, pre-training and transfer learning \cite{tlearn, tlearn1}. However, these strategies have many limitations, for example, they commonly rely heavily on the paired RGB data which is hard to acquire, and many of them can not be generalized to other tasks. 

Therefore, there is an urgent need for an efficient, event-data-specific data augmentation method that can be applied to various tasks. Our work aims to tackle this challenge within the event-based learning community, focusing on event data augmentation—a crucial technique for improving model generalization across different applications. We develop several effective data augmentation methods to increase the diversity of event datasets, overcoming the shortcomings of current techniques.
\subsection{Event Data Augmentation}

Data augmentation always plays an important role in enhancing the models' generalization ability. Former studies have proved that data augmentation is a practical technique for many different tasks \cite{da,da1,da2}. But for event data augmentation, only a small amount of work exists. Representative work includes NDA \cite{NDA}, EventDrop \cite{EventDrop} and EventMix \cite{EventMix}. NDA \cite{NDA}, which simply applies the data augmentation methods in RGB modality onto event like CutMix, flip and so on, does not provide a specific design that considers the unique sparse and spatio-temporal nature of event data, thus does not achieve a satisfactory result. EventDrop \cite{EventDrop}, which designs 3 kinds of random dropout strategies to improve the diversity of original datasets, and EventMix \cite{EventMix}, which applies a three-dimensional version of Mixup \cite{mixup} and CutMix \cite{cutmix} on the event data. Although they take both the spatial and temporal dimension into account, their augmentation performance is also unsatisfactory and unstable since they relies heavily on random and prior assumptions, ignoring the uneven and diverse distributions of different event data. Moreover, they only make preliminary use of event spatio-temporal information, thus failing to fully utilize the rich spatio-temporal relations inside event stream and are not robust to complex real world scenarios. 

In contrast, EventAug is tailored for event data, leveraging its unique properties. MSTI introduces spatio-temporal diversity through multi-scale integration, enhancing motion and object feature learning, as well as robustness to varying motion speeds. SSEM and TSEM enrich event data's high-level semantics by diversifying local spatial and temporal correlations, mitigating occlusion and motion disruption effects. Furthermore, guided by spatio-temporal saliency, our methods tackle uneven data distribution, ensuring efficiency and adaptability across diverse datasets.
\section{Method}\label{method}
\subsection{Overview}
The focus of EventAug is to simultaneously enhance the diversity of both the temporal and spatial dimensions, while also considering the uneven distribution of event data and the complexities of real-world scenarios. Therefore, we have designed a systematic enhancement framework that takes into account both temporal and spatial diversity, and increases the diversity in aspects such as occlusion and moving speed. Specifically, this includes three methods: \textit{Multi-scale Temporal Integration(MSTI)}, \textit{Spatial-salient Event Mask(SSEM)} and \textit{Temporal-salient Event Mask(TSEM)}
\subsection{Multi-scale Temporal Event Integration}
\subsubsection{Event Frame Integration}
State-of-the-art ANNs mainly deal with RGB frames, and can not directly process the sparse event streams. And for SNNs, directly inputting frames without preliminary encoding process has become a widely adopted strategy in deep spiking neural networks \cite{spikingjelly}. Therefore, to apply the existing powerful ANN and SNN models to the event vision and extract discriminative spatial cues, a mainstream solution is to transform event streams to frame-like data\cite{frame1,frame2,frame3,spikingjelly,eventframe}. We will introduce the details in the following:\\ 
Let \(E\) denotes the sequence of an event stream:
\begin{equation}
	E_i = (x_i,y_i,p_i,t_i)
\end{equation}
\((x_i,y_i)\) is the coordinate where the event \(E_i\) generates, \(t_i\) is the timestamp indicates when the event is generated, and \(p_i\) is the polarity with 1 and -1 indicating positive and negative events respectively. We pre-arrange the event stream in timestamp order. 

For the integration, we evenly divide the event stream into \(T\) slices. Let \(F(j)\) denotes the frame that generates from the \(j\)th event slice, \(i_{start}\) and \(i_{end}\) as the start and end timestamp of an event frame. So we have:
\begin{gather}
	i_{start} = \lfloor\frac{N}{T}\rfloor j\\
	i_{end} = \lfloor\frac{N}{T}\rfloor (j+1)
\end{gather}
Then, we perform temporal integration in the target time region:
\begin{gather}
	F(j)_{x_i,y_i,p_i} = \sum_{k={i_{start}}}^{i_{end}}\mathbb{I}(E_k)\\
	\mathbb{I}(E_i)=
	\begin{cases}
		1,& x_i = x_k, y_i = y_k, p_i = p_k\\
		0,& otherwise
	\end{cases}
\end{gather}
where \(\mathbb{I}\) is the indicator function, \(N\) is the total number of event in the event stream. After integration, each event stream transforms into \(T\) event frames, and each event frame can be treated as a 2 channel image with a resolution of [\(W\), \(H\)]. 
\subsubsection{Multi-scale Integration}\label{1.2.3}
This method is inspired by our observations that motion speed determines the completeness of motion cues and the clarity of object boundaries within an event frame. For a long-term temporal scale integration, more motion information is revealed including moving orbit, the speed of movement and so on. For a short-term temporal scale of integration, more information about the object itself is revealed(\textit{e.g.,} contours, shapes). These can be easily discerned from the visualization (Figure \ref{vis}).
Also, frames with different integration scales contain varying degrees of noise which is beneficial for feature extraction as diversity is improved. 
Based on our observation above, we design the multi-scale temporal integration to let neural network learn different kinds of pattern. We can also add diversity of motion speed to eliminate the negative effects brought by different moving speed and make the network more robust. In implementation, we apply a speed-aware policy: Choose both \(\frac{1}{n}\) and the m-fold of the base scale (n and m are hyper parameters), together with the base scale. This augmentation policy is highly general and can be applied to all datasets. Moreover, we find in experiments that we can achieve a great augmentation performance by simply setting hyper parameters to double and half scale without carefully tuning. The experimental results and analysis can be found in section \ref{experiment}.
\subsection{Spatial and Temporal Salient Event Mask}
To enrich local spatial and temporal correlation and improve robustness to occlusion and motion disruption, we propose two saliency-guided spatial and temporal masking method, namely \textit{Spatial-salient Event Mask(SSEM)} and \textit{Temporal-salient Event Mask(TSEM)}. Concretely, We first calculate the spatio-temporal saliency based on the distribution of event density in temporal and spatial patches, and then selectively apply masking on the temporal and spatial dimensions based on the saliency information.


\subsubsection{Spatial-salient Event Mask}
Considering the uneven spatial distribution of event data, we aim to improve the efficiency of our augmentation methods through guidance from saliency information. We first obtain the spatial saliency of the event data by a fast and training-free method we propose bellow. This method utilize the unique sparse nature of event data,which is different from the dense RGB modal images. Therefore, we can acquire the spatial saliency map of the event frame by observing the density distribution of it. With the sparse nature of event, the density distribution for event frame is a great approximation of event saliency map with very low computation cost since there is no need for training. 
To elaborate, for spatial-saliency, we first divide the event frame into \(16 \times 16\) patches like in \cite{MAE,vit}, then we obtain the saliency information by calculating the density distribution of event. Our detailed methods are described in Algorithm \ref{ssa}.
\begin{algorithm}
	\caption{Spatial-saliency calculation algorithm}
	\label{ssa}
		\begin{algorithmic}
			\Require \(E\) for original event stream, \(P_i\) for patch \(i\), one event frame have \(k\) patches in total. \(S\) denotes the patch saliency for each patch.
			\Ensure  \(idx\) (the salient patches indexes)
			\State \textbf{Init}: S = [], idx = [0,1,$\cdots$,k-1]
			\Function{spatial-saliency calculation}{}
			\For{\(P_i \in Patch\)}
			\For{\(E_j \in E\)}
			\If{\(E_j \in P_i\)}
			\State \(index\gets j\)
			\EndIf
			\EndFor
			\State \(S[i]\gets (len(index))\)
			\EndFor
			\State \Call{Sort}{\(idx[i]\) based on \(S[idx[i]]\) in descending order}
			\EndFunction
		\end{algorithmic}
\end{algorithm}
Given the spatial saliency information, we choose the most salient area of each frame to apply event spatial mask, which mask out all the events in the target patches. First, we define \(r\) as the mask rate. And we set a saliency threshold \(\epsilon\) to adjust the mask rate of the frame. \(\epsilon\) is decided by the density of the target event stream and mask rate \(r\), which ensure a percentage of \(r\) patches are masked. Let \(\mathbf{M}\) denotes the spatial-salient mask of a single event stream, \(F_o\) is the original event frame, \(F_M\) is the frame after augmentation,\(p\) denotes all the patches in \(F\), \(p_{s}\) is the salient patches we get from the algorithm above, function \(Dense(p_i)\) returns the event density of area \(p_i\) by performing the algorithm above. The detailed calculation method is as follows:

We first calculate the threshold of event density saliency \(\epsilon\):
\begin{gather}
	Idx = idx[kr-1]\\
	\epsilon = Dense(x_{Idx})
\end{gather}
Then, we obtain the spatial-salient mask of event frames by determining whether each patch is salient:
\begin{gather}
	\begin{cases}
		p_i \in p_s,& Dense(p_i)>\epsilon\\
		p_i \notin p_s,& otherwise
	\end{cases}\\
	\mathbf{M_{i,j}} = 		
	\begin{cases}
		0,& (i,j) \in p_{s}\\
		1,& otherwise
	\end{cases}
\end{gather}
Finally, we acquire the masked frame \(F_M\) by applying the Hadamard product of original frame \(F_o\) and mask \(\textbf{M}\).
\begin{gather}
	F_M = F_o \odot \mathbf{M}
\end{gather}
Now we get the masked frame \(F_M\).
\subsubsection{Temporal-salient Event Mask}
To address the uneven temporal distribution of event data, we propose \textit{Temporal-salient Event Mask}. We first calculate the temporal saliency of the event data by our proposed algorithm bellow. Just like Algorithm \ref{ssa}, this method utilizes the sparse nature of event data, thus it is fast and training-free Therefore, we can acquire the temporal saliency map of the event frame by observing its density distribution in temporal dimension. We first divide the event stream into \(T\) slices, where \(T\) is decided by the policies in \ref{1.2.3}. Detailed method is described in the left side of Algorithm \ref{tsa}.
\begin{algorithm*}
	\begin{minipage}{0.45\textwidth}
		\caption{Temporal-salient Event Mask}
		\label{tsa}
		\centering
		\begin{algorithmic}[1]
			\Require \(E\) for original event stream,\\ \(slice_i = [t_{start},t_{end}]\) for \\the timestamp of the \(i\)th event frame
			\Ensure  idx(the salient slices indexes)
			\State \textbf{Init}: idx = [0, 1,$\cdots$, T-1], count = 0
			\Function{temporal-saliency calculation}{}
			\For{\(slice_i \in slice\)}
			\State \(count = 0\)
			\For{\(E_j \in E\)}
			\If{\(E_j \in slice_i\)}
			\State \(count\gets count + 1\)
			\EndIf
			\EndFor
			\State \(s[i]\gets count\)
			\EndFor
			\State \Call{Sort}{\(idx[i]\) based on \(s[idx[i]]\) in descending order}
			\EndFunction
		\end{algorithmic}
	\end{minipage}
	\hfill
	\begin{minipage}{0.45\textwidth}
		\label{tmask}
		\begin{algorithmic}[1]
			\Require \(E\) for original event stream, \\\(slice\) for the target salient event frame slice,\\ \(p\) for a base mask rate, \(d = [d_1,d_2,\cdots, d_T]\) \\for the event density of the salient slice
			\Ensure  \(E_{m}\)(event stream after masking)
			\State \textbf{Init}: idx = [0, 1,$\cdots$, T], \(index_M\) = []
			\Function{temporal-salient event mask}{}
			\State \(m = min(d)\)
			\For{\(i = 0\) to \(T-1\)}
			\State \(p_s = d_i / m * p\)
			\State\(index = j\) if \(E_j \in slice_i\)
			\For{\(k\) in \(index\)}
			\If{\Call{random}{}(0,1) < \(p_s\)}
			\State \(index_m \gets k \)
			\EndIf
			\EndFor
			\State \(index_M = index_m \cup index_M\)
			\EndFor
			\State \(E_m = E \setminus E_{index_M}\)
			\EndFunction
		\end{algorithmic}
	\end{minipage}
\end{algorithm*}
For temporal-salient event mask, we first get the temporal saliency of an event stream by the algorithm above, then we apply the temporal-salient mask to let the network learn the rich spatio-temporal information of the event data better. Inside the salient frame slice, we first get the minimum density of the target salient frame slice \(m\), and we define a base mask rate \(p\) which is a hyper parameter. The mask rate \(p_s\) of each target salient slice will be decided by their event density and \(p\). For every event \(e\) in the slice, the mask probability of it is equal to \(p_s\). We describe this method in the right side of Algorithm \ref{tsa} in detail.
\section{Experiment}\label{experiment}
\subsection{Experiment Setup}
Our experiments are conducted using Pytorch \cite{Pytorch}, with Adam \cite{adam} optimizer and a learning rate of 0.001. For convergence, we train SNNs for 80 epochs and ANNs for 200 epochs. Detailed implementation information is provided in the supplementary materials.
To assess the generalization of our methods, we evaluate the augmentation technique on two distinct deep neural network architectures:
\begin{itemize}[leftmargin=*]
	\item \textbf{Spiking Neural Network (SNN)}: SNNs, due to their event-driven computing and temporal coding, are considered the most suitable network architecture for processing event data. Therefore, we choose the convolution spiking neural network (CSNN) defined and implemented by \cite{spikingjelly} as the backbone for experiment, which is a simple  SNN with 5 convolution layers and 3 full connection layers. The scale of the parameters is 1.7M. 
\item \textbf{Artificial Neural Network(ANN)}: We follow former studies \cite{est,EventDrop} to use Resnet-34 \cite{resnet34} as the backbone, which contains 4 residential layers for feature extraction. The scale of the parameters is 21.8M.
\end{itemize}
\subsection{Datasets}
We follow previous works \cite{EventMix,NDA,EventDrop} to use two challenging public datasets CIFAR10-DVS \cite{CIFAR10} and DVS128 Gesture \cite{DVSGesture} for evaluation. For CIFAR10-DVS, we follow \cite{EventMix,NDA,spikingjelly} to divide the training and test sets by 9 : 1 (9k train samples and 1k validation samples). DVS128 Gesture is a real-world gesture recognition dataset collected by the DVS camera. We follow \cite{EventMix,spikingjelly,SNN} to divide the training and test sets by 8:2 (1176 train samples and 288 validation samples).   
\subsection{Efficacy of our EventAug}

Table \ref{table1} and \ref{table2} compares our EventAug with other state-of-the-art event augmentation methods across different backbone architectures, using identical hyperparameters. EventAug consistently achieves significant performance improvements, showcasing its efficacy in enriching data diversity and reducing over-fitting.

NDA\cite{NDA} naively transfers RGB data augmentation to event data, neglecting the sparse and spatio-temporal aspects of event streams, leading to limited augmentation effects. EventDrop\cite{EventDrop} relies heavily on randomness and fails to address the uneven distribution of events, often resulting in the loss of crucial information or ineffective augmentation in irrelevant areas. In contrast, EventAug utilizes the sparse and spatio-temporal nature of event data, significantly boosting dataset diversity by capturing a broader spectrum of motion patterns and spatial-temporal correlations. This tailored approach enables EventAug to significantly outperform existing augmentation methods.

\begin{table}
	\caption{Classification accuracy (\%) of different networks with various augmentation techniques on CIFAR10-DVS dataset.}
	\label{table1}
	\centering
	\begin{tabular}{lll}
			\toprule
			\textbf{Model}     & \textbf{Method}     &  \textbf{Accuracy(Improvement)} \\
			\midrule
			\multirow{1}{*}{Resnet-34}  & Identity  &   74.20(+0.00)  \\
			& NDA \cite{NDA}&72.20(-2.00) \\
			&EventDrop\cite{EventDrop}&69.30(-4.90)\\
			&EventAug(Ours)&\textbf{75.60(+1.40)}\\
			\midrule
			\multirow{1}{*}{CSNN}  & Identity  & 74.80(+0.00)     \\
			& NDA \cite{NDA}& 76.50(+1.70)\\
			&EventDrop\cite{EventDrop}&75.80(+1.00)\\
			&EventAug(Ours)&\textbf{78.10(+3.30)}\\
			\bottomrule
		\end{tabular}
\end{table}
\begin{table*}
	\caption{Recognition accuracy (\%) of different networks with various state-of-the-art augmentation techniques on DVS128 Gesture dataset. \(^*\) means reference of original paper without running in our own environment.}
	\label{table2}
	\centering
	\begin{tabular}{lll}
			\toprule
			\textbf{Model}     & \textbf{Method}     &  \textbf{Accuracy(Improvement)} \\
                 \midrule
			\multirow{1}{*}{Resnet-34 (ANN)}  & Identity  & 95.49 (+0.00)  \\
			& NDA \cite{NDA}& 97.22(+1.73)\\
            &EventMix \cite{EventMix}&91.80\(^*\) (-3.69)\\
            &ShapeAug \cite{ShapeAug}&91.70\(^*\)(-3.79)\\	 
            &EventDrop\cite{EventDrop}&96.18(+0.69)\\
			&EventAug(Ours)&\textbf{98.26(+2.77)}\\
                \midrule
			\multirow{1}{*}{Resnet-18 (SNN)} & Identity  & 94.33\(^*\) (+0.00)  \\ 
   & EventMix\cite{EventMix}  & 96.75\(^*\)(+2.42)  \\
			& EventRPG\cite{eventrpg}& 96.53\(^*\)(+2.20)\\
			\midrule
			\multirow{1}{*}{CSNN (SNN)}  & Identity  & 93.75(+0.00)   \\
			& NDA \cite{NDA}&95.83(+2.08) \\
   &EventAugmentation \cite{eventaugmentation}&96.25\(^*\)(+2.50)\\
	   &EventDrop\cite{EventDrop}&94.44(+0.69)\\
			&EventAug(Ours)&\textbf{98.62(+4.87)}\\
			\bottomrule
		\end{tabular}
\end{table*}
\begin{table}
\caption{Accuracy (\%) of comparison of our proposed methods on different datasets with various models.}
\label{table3}
\centering
\begin{tabular}{llll}
	\toprule
	\multirow{2}{*}{\textbf{Model}}     & \multirow{2}{*}{\textbf{Method} }    &  \multicolumn{2}{c}{\textbf{Top-1 Accuracy (Improvement)} }\\
	\cline{3-4}
	&&CIFAR10-DVS&DVS128 Gesture\\
	\midrule
	\multirow{1}{*}{Resnet-34}  & Identity  & 74.20(+0.00)   &95.49(+0.00) \\
	& MSTI & 72.80(-1.40)&96.53(+1.04)\\ 
	&SSEM&75.60(+1.40)&97.92(+2.43)\\
	&TSEM&73.80(-0.40)&96.53(+1.04)\\
	\midrule
	\multirow{1}{*}{CSNN}  & Identity & 74.80(+0.00)  &93.75(+0.00) \\
	& MSTI &78.10(+3.30)&97.57(+3.82) \\
	&SSEM&76.40(+1.60)&96.18(+2.43)\\
	&TSEM&76.70(+1.90)&98.62(+4.87)\\
	\bottomrule
\end{tabular}
\end{table}
\subsection{Ablation Study} 
We performed ablative experiments on CIFAR10-DVS and DVS128 Gesture datasets using two backbones to evaluate our EventAug methods. The outcomes are summarized in Tables \ref{table1} and \ref{table2}.

Table \ref{table3} shows that our three augmentation methods significantly improve most of models and tasks, especially on SNNs. For instance, we achieved a 3.30\% accuracy gain on CIFAR10-DVS and 4.87\% gain on DVS128 Gesture. However, our method does not show significant improvement when tested on the CIFAR10-DVS dataset with Resnet-34. We believe that the reason for this is that the dataset is based on static images, and the details will be discussed in the \textbf{Limitation} section.

In Table \ref{table4}, we observe that ResNet-34, with its focus on spatial information, benefits from spatial-oriented augmentations, such as SSEM, which boosts accuracy to 97.92\% (+2.43). When all enhancement methods are combined, accuracy further improves to 98.26\% (+2.77), indicating that enriching spatio-temporal diversity in the dataset yields significant performance gains.

For Table \ref{t3}, we conduct experiments on various scales to prove that the benefits of the multi-scale strategy come from the integration of multiple scales, not just the optimal scale. We believe that the poor performance at short scales is due to the lack of object information at long scales, making it difficult for the model to accurately learn the motion semantics. Therefore, our MSTI approach, which incorporates both short and long temporal scales, utilizes multi-scale temporal cues in a complementary manner and results in significant improvements.

We also show that adding saliency to the masking strategy enhances augmentation in appendix.

\begin{figure}[h]
\centering
\includegraphics[width=1.0\linewidth]{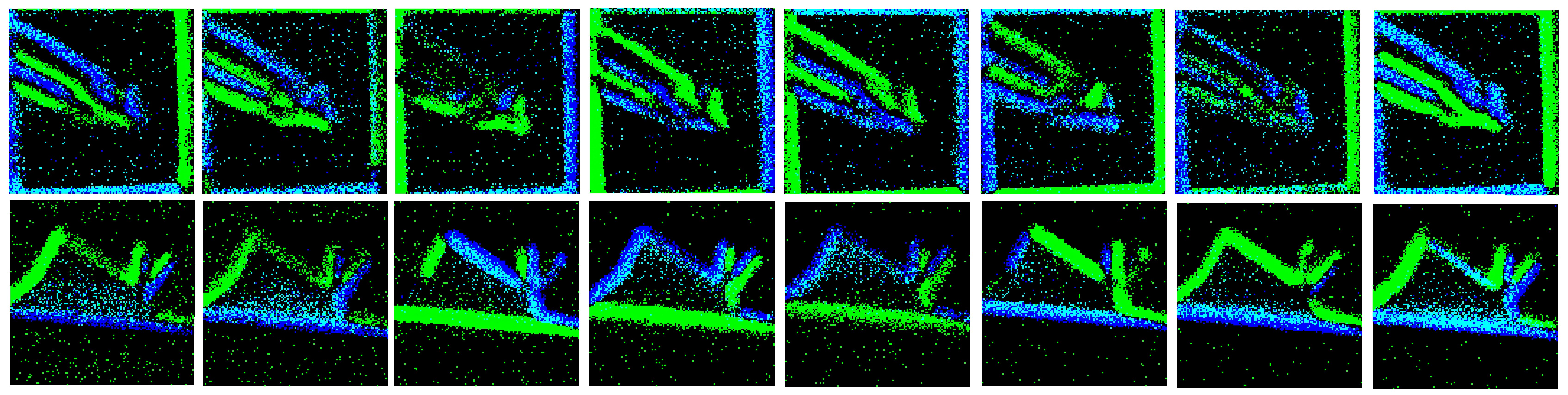}
\caption{Examples of consecutive frames in CIFAR10-DVS datasets.}
\label{limitationa}
\end{figure}
\begin{table}
\caption{Performance of EventAug for Resnet-34 on DVS128 Gesture with different augmentation setting(\%).}
\label{table4}
\centering
\begin{tabular}{lllll}
	\toprule
	\textbf{Model}   & \textbf{MSTI} &\textbf{SSEM}  & \textbf{TSEM}&\textbf{Top-1 Accuracy} \\
	\midrule
	\multirow{1}{*}{Resnet-34}  & \(\times\) & \(\times\) &\(\times\) & 95.49(+0.00)\\
	&\checkmark &\(\times\)&\(\times\)&96.53(+1.04)\\
	&\(\times\)&\checkmark&\(\times\)&97.92(+2.43)\\
	&\(\times\)&\(\times\)&\checkmark&96.53(+1.04)\\
	&\(\times\)&\checkmark&\checkmark&97.57(+2.08)\\ 
	&\checkmark&\checkmark&\checkmark&\textbf{98.26(+2.77)}\\
	\bottomrule
\end{tabular}
\end{table}

\begin{table}
\caption{Comparison of Accuracy (\%) of different integration scale for SNN on classification and recognition tasks.}
\label{t3}
\centering
\begin{tabular}{lll}
	\toprule
	\multirow{2}{*}{\textbf{Integration Scale} }    &  \multicolumn{2}{c}{\textbf{Top-1 Accuracy (Improvement)} }\\
	\cline{2-3}
	&CIFAR10-DVS&DVS128 Gesture\\
	\midrule
	Base scale  & 74.80(+0.00)   &93.75(+0.00) \\
	Short-term scale&73.70(-1.10) &93.40(-0.35)\\ 
	Long-term scale&76.60(+1.80)&95.83(+2.08)\\
	MSTI (ours)&\textbf{78.10(+3.30)}&\textbf{97.57 (+3.82)}\\
	\bottomrule
\end{tabular}
\end{table}

\section{Limitation}\label{limitation}
Our EventAug and prior methods yield sub-optimal CIFAR-10 classification with ResNet-34. We believe this is due to the dataset being derived from static images (CIFAR-10\cite{CIFAR}) rather than being directly captured from the real world. This leads to high similarity between consecutive frames (Figure \ref{limitationa}), emphasizing spatial information over temporal information. ResNet-34, unlike SNNs, emphasizes spatial embedding in event frames, neglecting temporal semantics, thus limiting the enhancement from our temporal augmentation techniques like MSTI.


\section{Conclusion}\label{conclusion}
In this work, we introduce a spatio-temporal data augmentation method that diversifies motion speeds and local correlations using three strategies. EventAug improves model robustness in challenging scenes and shows strong generalization across different network architectures. Our approach achieves significant improvements, as validated by experiments with multiple backbones and tasks. In the future, we will expand this augmentation method to other event-based learning tasks like detection, estimation and segmentation. 

\bibliography{aaai25}

\begin{thebibliography}{41}
\providecommand{\natexlab}[1]{#1}

\bibitem[{Amir et~al.(2017)Amir, Taba, Berg, Melano, McKinstry, Di~Nolfo, Nayak, Andreopoulos, Garreau, Mendoza, Kusnitz, Debole, Esser, Delbruck, Flickner, and Modha}]{DVSGesture}
Amir, A.; Taba, B.; Berg, D.; Melano, T.; McKinstry, J.; Di~Nolfo, C.; Nayak, T.; Andreopoulos, A.; Garreau, G.; Mendoza, M.; Kusnitz, J.; Debole, M.; Esser, S.; Delbruck, T.; Flickner, M.; and Modha, D. 2017.
\newblock A Low Power, Fully Event-Based Gesture Recognition System.
\newblock In \emph{2017 IEEE Conference on Computer Vision and Pattern Recognition (CVPR)}, 7388--7397.

\bibitem[{Bendig, Schuster, and Stricker(2024)}]{ShapeAug}
Bendig, K.; Schuster, R.; and Stricker, D. 2024.
\newblock ShapeAug: Occlusion Augmentation for Event Camera Data.
\newblock \emph{ArXiv}, abs/2401.02274.

\bibitem[{Cheng et~al.(2020)Cheng, Luo, Yang, Yu, and Li}]{CIFAR10}
Cheng, W.; Luo, H.; Yang, W.; Yu, L.; and Li, W. 2020.
\newblock Structure-Aware Network for Lane Marker Extraction with Dynamic Vision Sensor.
\newblock arXiv:2008.06204.

\bibitem[{Deng et~al.(2022)Deng, Chen, Liu, and Li}]{classification}
Deng, Y.; Chen, H.; Liu, H.; and Li, Y. 2022.
\newblock A Voxel Graph CNN for Object Classification with Event Cameras.
\newblock In \emph{2022 IEEE/CVF Conference on Computer Vision and Pattern Recognition (CVPR)}, 1162--1171.

\bibitem[{Dosovitskiy et~al.(2021)Dosovitskiy, Beyer, Kolesnikov, Weissenborn, Zhai, Unterthiner, Dehghani, Minderer, Heigold, Gelly, Uszkoreit, and Houlsby}]{vit}
Dosovitskiy, A.; Beyer, L.; Kolesnikov, A.; Weissenborn, D.; Zhai, X.; Unterthiner, T.; Dehghani, M.; Minderer, M.; Heigold, G.; Gelly, S.; Uszkoreit, J.; and Houlsby, N. 2021.
\newblock An Image is Worth 16x16 Words: Transformers for Image Recognition at Scale.
\newblock In \emph{9th International Conference on Learning Representations, {ICLR} 2021, Virtual Event, Austria, May 3-7, 2021}. OpenReview.net.

\bibitem[{Fang et~al.(2023)Fang, Chen, Ding, Yu, Masquelier, Chen, Huang, Zhou, Li, and Tian}]{spikingjelly}
Fang, W.; Chen, Y.; Ding, J.; Yu, Z.; Masquelier, T.; Chen, D.; Huang, L.; Zhou, H.; Li, G.; and Tian, Y. 2023.
\newblock SpikingJelly: An open-source machine learning infrastructure platform for spike-based intelligence.
\newblock arXiv:2310.16620.

\bibitem[{Fang et~al.(2021{\natexlab{a}})Fang, Yu, Chen, Huang, Masquelier, and Tian}]{frame2}
Fang, W.; Yu, Z.; Chen, Y.; Huang, T.; Masquelier, T.; and Tian, Y. 2021{\natexlab{a}}.
\newblock Deep Residual Learning in Spiking Neural Networks.
\newblock In \emph{Neural Information Processing Systems}.

\bibitem[{Fang et~al.(2021{\natexlab{b}})Fang, Yu, Chen, Masquelier, Huang, and Tian}]{SNN}
Fang, W.; Yu, Z.; Chen, Y.; Masquelier, T.; Huang, T.; and Tian, Y. 2021{\natexlab{b}}.
\newblock Incorporating Learnable Membrane Time Constant to Enhance Learning of Spiking Neural Networks.
\newblock In \emph{2021 IEEE/CVF International Conference on Computer Vision (ICCV)}, 2641--2651.

\bibitem[{Gehrig et~al.(2019)Gehrig, Loquercio, Derpanis, and Scaramuzza}]{est}
Gehrig, D.; Loquercio, A.; Derpanis, K.; and Scaramuzza, D. 2019.
\newblock End-to-End Learning of Representations for Asynchronous Event-Based Data.
\newblock In \emph{2019 IEEE/CVF International Conference on Computer Vision (ICCV)}, 5632--5642.

\bibitem[{Gu et~al.(2024)Gu, Dou, Li, Long, Guo, Chen, Liu, Jiao, and Li}]{eventaugmentation}
Gu, F.; Dou, J.; Li, M.; Long, X.; Guo, S.; Chen, C.; Liu, K.; Jiao, X.; and Li, R. 2024.
\newblock EventAugment: Learning Augmentation Policies From Asynchronous Event-Based Data.
\newblock \emph{IEEE Transactions on Cognitive and Developmental Systems}, 16(4): 1521--1532.

\bibitem[{Gu et~al.(2021)Gu, Sng, Hu, and Yu}]{EventDrop}
Gu, F.; Sng, W.; Hu, X.; and Yu, F. 2021.
\newblock EventDrop: Data Augmentation for Event-based Learning.
\newblock In Zhou, Z.-H., ed., \emph{Proceedings of the Thirtieth International Joint Conference on Artificial Intelligence, {IJCAI-21}}, 700--707. International Joint Conferences on Artificial Intelligence Organization.
\newblock Main Track.

\bibitem[{He et~al.(2022)He, Chen, Xie, Li, Doll\'ar, and Girshick}]{MAE}
He, K.; Chen, X.; Xie, S.; Li, Y.; Doll\'ar, P.; and Girshick, R. 2022.
\newblock Masked Autoencoders Are Scalable Vision Learners.
\newblock In \emph{Proceedings of the IEEE/CVF Conference on Computer Vision and Pattern Recognition (CVPR)}, 16000--16009.

\bibitem[{He et~al.(2016)He, Zhang, Ren, and Sun}]{resnet34}
He, K.; Zhang, X.; Ren, S.; and Sun, J. 2016.
\newblock Deep Residual Learning for Image Recognition.
\newblock In \emph{2016 IEEE Conference on Computer Vision and Pattern Recognition (CVPR)}, 770--778.

\bibitem[{Kaiser, Mostafa, and Neftci(2020)}]{frame1}
Kaiser, J.; Mostafa, H.; and Neftci, E. 2020.
\newblock Synaptic Plasticity Dynamics for Deep Continuous Local Learning (DECOLLE).
\newblock \emph{Frontiers in Neuroscience}, 14.

\bibitem[{Kingma and Ba(2017)}]{adam}
Kingma, D.~P.; and Ba, J. 2017.
\newblock Adam: A Method for Stochastic Optimization.
\newblock arXiv:1412.6980.

\bibitem[{Klenk et~al.(2022)Klenk, Bonello, Koestler, and Cremers}]{tlearn2}
Klenk, S.; Bonello, D.; Koestler, L.; and Cremers, D. 2022.
\newblock Masked Event Modeling: Self-Supervised Pretraining for Event Cameras.
\newblock \emph{2024 IEEE/CVF Winter Conference on Applications of Computer Vision (WACV)}, 2367--2377.

\bibitem[{Krizhevsky(2009)}]{CIFAR}
Krizhevsky, A. 2009.
\newblock Learning Multiple Layers of Features from Tiny Images.
\newblock \url{https://www.cs.toronto.edu/~kriz/cifar.html}.
\newblock (Updated 2019).

\bibitem[{Krizhevsky, Sutskever, and Hinton(2017)}]{da}
Krizhevsky, A.; Sutskever, I.; and Hinton, G.~E. 2017.
\newblock ImageNet classification with deep convolutional neural networks.
\newblock \emph{Commun. ACM}, 60(6): 84–90.

\bibitem[{Kumar et~al.(2023)Kumar, Mileo, Brennan, and Bendechache}]{da1}
Kumar, T.; Mileo, A.; Brennan, R.; and Bendechache, M. 2023.
\newblock Image Data Augmentation Approaches: A Comprehensive Survey and Future directions.

\bibitem[{Li et~al.(2022)Li, Kim, Park, Geller, and Panda}]{NDA}
Li, Y.; Kim, Y.; Park, H.; Geller, T.; and Panda, P. 2022.
\newblock Neuromorphic Data Augmentation for Training Spiking Neural Networks.
\newblock In Avidan, S.; Brostow, G.; Ciss{\'e}, M.; Farinella, G.~M.; and Hassner, T., eds., \emph{Computer Vision -- ECCV 2022}, 631--649. Cham: Springer Nature Switzerland.
\newblock ISBN 978-3-031-20071-7.

\bibitem[{Lichtsteiner, Posch, and Delbruck(2008{\natexlab{a}})}]{eventcamera}
Lichtsteiner, P.; Posch, C.; and Delbruck, T. 2008{\natexlab{a}}.
\newblock A 128$\times$ 128 120 dB 15 $\mu$s Latency Asynchronous Temporal Contrast Vision Sensor.
\newblock \emph{IEEE Journal of Solid-State Circuits}, 43(2): 566--576.

\bibitem[{Lichtsteiner, Posch, and Delbruck(2008{\natexlab{b}})}]{depthestimation}
Lichtsteiner, P.; Posch, C.; and Delbruck, T. 2008{\natexlab{b}}.
\newblock A 128$\times$ 128 120 dB 15 $\mu$s Latency Asynchronous Temporal Contrast Vision Sensor.
\newblock \emph{IEEE Journal of Solid-State Circuits}, 43(2): 566--576.

\bibitem[{Nazari and Faez(2019)}]{nc3}
Nazari, S.; and Faez, K. 2019.
\newblock Establishing the flow of information between two bio-inspired spiking neural networks.
\newblock \emph{Information Sciences}, 477: 80--99.

\bibitem[{Paszke et~al.(2017)Paszke, Gross, Chintala, Chanan, Yang, DeVito, Lin, Desmaison, Antiga, and Lerer}]{Pytorch}
Paszke, A.; Gross, S.; Chintala, S.; Chanan, G.; Yang, E.; DeVito, Z.; Lin, Z.; Desmaison, A.; Antiga, L.; and Lerer, A. 2017.
\newblock Automatic differentiation in PyTorch.
\newblock In \emph{NIPS-W}.

\bibitem[{Peng et~al.(2023)Peng, Zhang, Xiong, Sun, and Wu}]{GET}
Peng, Y.; Zhang, Y.; Xiong, Z.; Sun, X.; and Wu, F. 2023.
\newblock GET: Group Event Transformer for Event-Based Vision.
\newblock In \emph{2023 IEEE/CVF International Conference on Computer Vision (ICCV)}, 6015--6025.

\bibitem[{Ponghiran, Liyanagedera, and Roy(2023)}]{flowestimation}
Ponghiran, W.; Liyanagedera, C.~M.; and Roy, K. 2023.
\newblock Event-based Temporally Dense Optical Flow Estimation with Sequential Learning.
\newblock In \emph{2023 IEEE/CVF International Conference on Computer Vision (ICCV)}, 9793--9802.

\bibitem[{Rebecq, Horstschaefer, and Scaramuzza(2017)}]{eventframe}
Rebecq, H.; Horstschaefer, T.; and Scaramuzza, D. 2017.
\newblock Real-time Visual-Inertial Odometry for Event Cameras using Keyframe-based Nonlinear Optimization.
\newblock In \emph{British Machine Vision Conference}.

\bibitem[{Ren et~al.(2023)Ren, Zhou, Huang, Fu, Lin, Song, and Cheng}]{spikepoint}
Ren, H.; Zhou, Y.; Huang, Y.; Fu, H.; Lin, X.; Song, J.; and Cheng, B. 2023.
\newblock Spikepoint: An efficient point-based spiking neural network for event cameras action recognition.
\newblock \emph{arXiv preprint arXiv:2310.07189}.

\bibitem[{Saini and Malik(2021)}]{da2}
Saini, D.; and Malik, R. 2021.
\newblock Image Data Augmentation techniques for Deep Learning -A Mirror Review.
\newblock In \emph{2021 9th International Conference on Reliability, Infocom Technologies and Optimization (Trends and Future Directions) (ICRITO)}, 1--5.

\bibitem[{Shen, Zhao, and Zeng(2023)}]{EventMix}
Shen, G.; Zhao, D.; and Zeng, Y. 2023.
\newblock EventMix: An efficient data augmentation strategy for event-based learning.
\newblock \emph{Information Sciences}, 644: 119170.

\bibitem[{Stoffregen et~al.(2019)Stoffregen, Gallego, Drummond, Kleeman, and Scaramuzza}]{motionsegmentation}
Stoffregen, T.; Gallego, G.; Drummond, T.; Kleeman, L.; and Scaramuzza, D. 2019.
\newblock Event-Based Motion Segmentation by Motion Compensation.
\newblock In \emph{2019 IEEE/CVF International Conference on Computer Vision (ICCV)}, 7243--7252.

\bibitem[{Sun et~al.(2024)Sun, Zhang, Ge, Wang, Li, Fang, and Xu}]{eventrpg}
Sun, M.; Zhang, D.; Ge, Z.; Wang, J.; Li, J.; Fang, Z.; and Xu, R. 2024.
\newblock EventRPG: Event Data Augmentation with Relevance Propagation Guidance.
\newblock In \emph{The Twelfth International Conference on Learning Representations, {ICLR} 2024, Vienna, Austria, May 7-11, 2024}. OpenReview.net.

\bibitem[{Wang et~al.(2023)Wang, Weng, Zhang, and Xiong}]{derain}
Wang, J.; Weng, W.; Zhang, Y.; and Xiong, Z. 2023.
\newblock Unsupervised Video Deraining with An Event Camera.
\newblock In \emph{2023 IEEE/CVF International Conference on Computer Vision (ICCV)}, 10797--10806.

\bibitem[{Wang, Chae, and Yoon(2021)}]{tlearn}
Wang, L.; Chae, Y.; and Yoon, K.-J. 2021.
\newblock Dual Transfer Learning for Event-based End-task Prediction via Pluggable Event to Image Translation.
\newblock In \emph{2021 IEEE/CVF International Conference on Computer Vision (ICCV)}, 2115--2125.

\bibitem[{Wu et~al.(2019)Wu, Deng, Li, Zhu, Xie, and Shi}]{frame3}
Wu, Y.; Deng, L.; Li, G.; Zhu, J.; Xie, Y.; and Shi, L. 2019.
\newblock Direct training for spiking neural networks: faster, larger, better.
\newblock In \emph{Proceedings of the Thirty-Third AAAI Conference on Artificial Intelligence and Thirty-First Innovative Applications of Artificial Intelligence Conference and Ninth AAAI Symposium on Educational Advances in Artificial Intelligence}, AAAI'19/IAAI'19/EAAI'19. AAAI Press.
\newblock ISBN 978-1-57735-809-1.

\bibitem[{Yang, Pan, and Liu(2023)}]{tlearn1}
Yang, Y.; Pan, L.; and Liu, L. 2023.
\newblock Event Camera Data Pre-training.
\newblock In \emph{2023 IEEE/CVF International Conference on Computer Vision (ICCV)}, 10665--10675.

\bibitem[{Yun et~al.(2019)Yun, Han, Chun, Oh, Yoo, and Choe}]{cutmix}
Yun, S.; Han, D.; Chun, S.; Oh, S.~J.; Yoo, Y.; and Choe, J. 2019.
\newblock CutMix: Regularization Strategy to Train Strong Classifiers With Localizable Features.
\newblock In \emph{2019 IEEE/CVF International Conference on Computer Vision (ICCV)}, 6022--6031.

\bibitem[{Zhang et~al.(2018)Zhang, Cisse, Dauphin, and Lopez-Paz}]{mixup}
Zhang, H.; Cisse, M.; Dauphin, Y.~N.; and Lopez-Paz, D. 2018.
\newblock mixup: Beyond Empirical Risk Minimization.
\newblock arXiv:1710.09412.

\bibitem[{Zhang et~al.(2023)Zhang, Yu, Yang, Liu, and Xia}]{motiondeblur}
Zhang, X.; Yu, L.; Yang, W.; Liu, J.; and Xia, G.-S. 2023.
\newblock Generalizing Event-Based Motion Deblurring in Real-World Scenarios.
\newblock In \emph{2023 IEEE/CVF International Conference on Computer Vision (ICCV)}, 10700--10710.

\bibitem[{Zhao et~al.(2022)Zhao, Li, Zeng, Wang, and Zhang}]{nc2}
Zhao, D.; Li, Y.; Zeng, Y.; Wang, J.; and Zhang, Q. 2022.
\newblock Spiking CapsNet: A spiking neural network with a biologically plausible routing rule between capsules.
\newblock \emph{Information Sciences}, 610: 1--13.

\bibitem[{Zhao et~al.(2020)Zhao, Zeng, Zhang, Shi, and Zhao}]{nc1}
Zhao, D.; Zeng, Y.; Zhang, T.; Shi, M.; and Zhao, F. 2020.
\newblock GLSNN: A Multi-Layer Spiking Neural Network Based on Global Feedback Alignment and Local STDP Plasticity.
\newblock \emph{Frontiers in Computational Neuroscience}, 14.

\end{thebibliography}

\end{document}